%% file: 2018_wmt_cognate.tex
\pdfoutput=1
\documentclass[11pt,a4paper]{article}
\PassOptionsToPackage{hyphens}{url}
\usepackage[hyperref]{emnlp2018}
\usepackage{times}
\usepackage{latexsym}

\usepackage[utf8]{inputenc}
\usepackage[T1]{fontenc}
\usepackage[hyphens]{url}
\usepackage{booktabs}
\usepackage{graphicx}   

\aclfinalcopy 

\input{core.tex}
\DeclareMathOperator*{\softmax}{softmax}
\DeclareMathOperator*{\attention}{Att}
\DeclareMathOperator*{\multihead}{MH}
\DeclareMathOperator*{\feedforward}{FF}

\title{Cognate-aware morphological segmentation for multilingual neural translation}

\author{Stig-Arne Grönroos \\
  {\tt stig-arne.gronroos@aalto.fi} \\
  Aalto University, Finland \\\And
  Sami Virpioja \\
  {\tt sami.virpioja@aalto.fi} \\
  Aalto University, Finland \\
  Utopia Analytics, Finland \\\AND
  Mikko Kurimo \\
  {\tt mikko.kurimo@aalto.fi} \\
  Aalto University, Finland}

\date{}

\begin{document}
\maketitle

\begin{abstract}
This article describes the Aalto University entry to the WMT18 News Translation Shared Task.
We participate in the multilingual subtrack
with a system trained under the constrained condition to translate from English to both Finnish and Estonian.
The system is based on the Transformer model.
We focus on improving the consistency of morphological segmentation for words that are
similar orthographically, semantically, and distributionally;
such words include etymological cognates, loan words, and proper names.
For this, we introduce Cognate Morfessor, a multilingual variant of the Morfessor method.
We show that our approach improves the translation quality particularly for Estonian,
which has less resources for training the translation model.
\end{abstract}

\section{Introduction}

\begin{table*}
\begin{center}
{\small
\begin{tabular}{lllll}
\toprule
type    & consistent & en                       & fi                        & et \\
\midrule
(i)     & yes        & On + y + sz + kie + wicz & On + y + sz + kie + wicz  & On + y + sz + kie + wicz  \\
(ii)    & yes        & gett + ing               & saa + mise + ksi          & saa + mise + ks           \\
        &            & work + ing               & toimi + mise + ksi        & toimi + mise + ks         \\
(iii)   & yes        & work time                & työ + aja + sta           & töö + aja + st            \\
\midrule
(i)     & no         & On + y + sz + kie + wicz & Onys + zk + ie + wi + cz  & O + nysz + ki + ewicz     \\
(ii)    & no         & get + ting               & saami + seksi             & saami + seks              \\
        &            & work + ing               & toimi + mise + ksi        & toimi + miseks            \\
(iii)   & no         & work time                & työ + aja + sta           & tööajast                  \\
\bottomrule
\end{tabular}
}
\caption{Example consistent and inconsistent segmentations.
    \label{tab:consistent}}
\end{center}
\end{table*}

Cognates are words in different languages,
which due to a shared etymological origin
are represented as identical or nearly identical strings,
and also refer to the same or similar concepts.
Ideally the cognate pair is similar orthographically, semantically, and distributionally.
Care must be taken with ``false friends'',
i.e. words with similar string representation but different semantics.
Following usage in Natural Language Processing, e.g. \cite{kondrak2001identifying},
we use this broader definition of the term cognate,
without placing the same weight on etymological origin as in historical linguistics.
Therefore we accept loan words as cognates.

In any language pair written in the same alphabet,
cognates can be found among names of persons, locations and other proper names.
Cognates are more frequent in related languages, such as Finnish and Estonian.
These additional cognates are words of any part-of-speech,
which happen to have a shared origin.

In this work we set out to improve morphological segmentation for multilingual translation systems
with one source language and two related target languages.
One of the target languages is assumed to be a low-resource language.
The motivation for using such a system is to exploit the large resources of a related language
in order to improve the quality of translation into the low-resource language.

Consistency of the segmentations is important when using subword units
in machine translation.  We identify three types of consistency in the
multilingual translation setting (see examples in
Table~\ref{tab:consistent}):

(i) The benefit of consistency is most evident when the translated word is an identical cognate
between the source and a target language.
If the source and target segmentations are consistent,
such words can be translated by sequentially copying subwords from source to target.

(ii) Language-internal consistency means
that when a subword boundary is added, its location corresponds to a true morpheme boundary,
and that if some morpheme boundaries are left unsegmented, the choices are consistent between words.
This improves the productivity of the subwords
and reduces the risk of introducing short, word-internal errors at the subword boundaries.
In the example \examp{*saami + miseksi},
choosing the wrong second morph causes the letters \emph{mi} to be accidentally repeated.

(iii) When training a multilingual model,
a third form of consistency arises between the different target languages.
An optimal segmentation would maximize the use of morphemes
with cross-lingually similar string representations and meanings,
whether they occur in cognate words or elsewhere.
We hypothesize that segmentation consistency between target languages
enables learning of better generalizing subword representations.
This consistency allows contexts seen in the high-resource corpus
to fill in for those  missing from the low-resource corpus.
This should lead to improved translation results, especially for the lower resourced target language.

Naïve joint training of a segmentation model,
e.g. by training Byte Pair Encoding (BPE) \cite{sennrich2015neural}
on the concatenation of the training corpora in different languages,
can only address consistency when the cognates are identical (type \emph{i}),
or with some luck if the differences occur in the ends of the words.
If a single letter changes in the middle of a cognate, consistent subwords that span
over the location of the change are found only by chance.
In order to encourage stronger consistency,
we propose a segmentation model that uses automatically extracted cognates
and fuzzy matching between cognate morphs.

In this work we also contribute two new features to the OpenNMT translation system:
Ensemble decoding, and fine-tuning a pre-trained model using a compatible data set.%
\footnote{Our changes are awaiting inclusion in OpenNMT.
In the mean time, they are available from \url{https://github.com/Waino/OpenNMT-py/tree/ensemble}}

\subsection{Related work}
Improving segmentation through multilingual learning has been studied before.
\newcite{snyder2008unsupervised} propose an unsupervised, Bayesian method,
which only uses parallel phrases as training data.
\newcite{wicentowski2004multilingual} present a supervised method, which requires lemmatization.
The method of \newcite{naradowsky2011unsupervised} is also unsupervised, 
utilizing a hidden semi-Markov model, but it requires rich features on the input data.

The subtask of cognate extraction has seen much research effort
\cite{mitkov2007methods,bloodgood2017using,ciobanu2014automatic}.
Most methods are supervised, and/or require rich features.

There is also work on cognate identification from historical linguistics perspective
\cite{rama2016siamese,kondrak2009identification},
where the aim is to classify which cognate candidates truly share an etymological origin.

We propose a language-agnostic, unsupervised method,
which doesn't require annotations, lemmatizers, analyzers or parsers.
Our method can exploit both monolingual and parallel data,
and can use cognates of any part-of-speech.

\section{Cognate Morfessor}

We introduce a new variant of Morfessor for cross-lingual segmentation.%
\footnote{Available from \url{https://github.com/Waino/morfessor-cognates}}
It is trained using a bilingual corpus,
so that both target languages are trained simultaneously.

We allow each language to have its own subword lexicon.
In essence,
as a Morfessor model consists of a lexicon and the corpus encoded with that lexicon,
we now have two separate complete Morfessor sub-models. 
The two models are linked through the training algorithm.
We want the segmentation of non-cognates to tend towards the normal Morfessor Baseline segmentation,
but place some additional constraints on how the cognates are segmented.

In our first experiments,
we only restricted the number of subwords on both sides of the cognate pair to be equal.
This criterion was too loose,
and we saw many of the longer cognates segmented with both 1-to-N and N-to-1 morpheme correspondences.
For example
\begin{center}
\begin{tabular}{llllllll}
ty  &+& ö   &+& aja &+& sta \\
töö &+& aja &+& s   &+& t   \\
\end{tabular}
\end{center}

To further encourage consistency,
we included a third component to the model,
which encodes the letter edits transforming the subwords of one cognate into the other.

Cognate Morfessor is inspired by Allomorfessor \cite{kohonen09clef_lncs,virpioja2010lncs},
which is a variant of Morfessor that includes modeling of allomorphic variation.
Simultaneously to learning the segmentations, 
Allomorfessor learns a lexicon of transformations to convert a morph into one of its allomorphs.
Allomorfessor is trained on monolingual data.

We implement the new version as an extension of Morfessor Baseline 2.0 \cite{virpioja2013tr}.

\subsection{Model}
The Morfessor Baseline cost function \cite{creutz02sigphon}
\begin{align}
\cost(\params, \data) = -\log p(\params) - \log p(\data \vb \params)
\end{align}
is extended to
\begin{align}
\cost(\params, \data) =
    & -\log p(\params_{1}) 
      -\log p(\params_{2}) 
      -\log p(\params_{E}) \nonumber\\
    & - \log p(\data_{1} \vb \params_{1})
      - \log p(\data_{2} \vb \params_{2}) \nonumber\\
    & - \log p(\data_{E} \vb \params_{E}) \label{eq:cost}
\end{align}
dividing both lexicon and corpus coding costs into three parts:
one for each language ($\params_{1},\data_{1}$ and $\params_{2},\data_{2}$)
and one for the edits transforming the cognates from one language to the other ($\params_{E},\data_{E}$).

The coding is redundant, as one language and the edits would be enough to reconstruct the second language.
In the interest of symmetry between target languages, we ignore this redundancy.

The intuition is that the changes in spelling between the cognates in a particular language pair is regular.
Coding the differences in a way that reduces the cost of making a similar change
in another word guides the model towards learning these patterns from the data.

The coding of the edits is based on the \newcite{levenshtein1966} algorithm.
Let $(w^{a}, w^{b})$ be a cognate pair 
and its current segmentation $\big((m^{a}_{1}, \mathellipsis, m^{a}_{n}), (m^{b}_{1}, \mathellipsis m^{b}_{n})\big)$.
The morphs are paired up sequentially.
Note that the restrictions on the search algorithm guarantee that 
both segmentations contain the same number of morphs, $n$.
For a morph pair $(m^{a}_{i}, m^{b}_{i})$, the Levenshtein-minimal set of edits is calculated.
Edits that are immediately adjacent to each other are merged.
In order to improve the modeling of sound length change,
we extend the edit in both languages to include the neighboring unchanged character,
if one half of the edit is the empty string $\epsilon$,
and the other contains another instance of character representing the sound being lengthened or shortened.
This extension encodes a sound lengthening as e.g. 'a$\rightarrow$aa'
instead of '$\epsilon\rightarrow$a'.
As the edits are cheaper to reuse once added to the edit lexicon,
avoiding edits with $\epsilon$ on either side is beneficial to reduce spurious use.
Finally, position information is discarded from the edits, leaving only the substrings,
separated by a boundary symbol.

As an example, the edits found between 
\examp{yhteenkuuluvuuspolitiikkaa} and \examp{ühtekuuluvuspoliitika}
are 'y$\rightarrow$ü', 'een$\rightarrow$e', 'uu$\rightarrow$u', 'ti$\rightarrow$it', and 'kka$\rightarrow$k'.

The semi-supervised weighting scheme of \newcite{kohonen2010sigmorphon}
can be applied to Cognate Morfessor.
A new weighting parameter \emph{edit\_cost\_weight} is added,
and multiplicatively applied to both the lexicon and corpus costs of the edits.

The training algorithm is an iterative greedy local search very similar to the Morfessor Baseline algorithm.
The algorithm finds an approximately minimizing solution to Eq \ref{eq:cost}.
The recursive splitting algorithm from Morfessor Baseline is slightly modified.
If a non-cognate is being reanalyzed, the normal algorithm is followed.
Cognates are reanalyzed together.
Recursive splitting is applied,
with the restriction that if a morph in one language is split,
then the corresponding cognate morph in the other language must be split as well.
The Cartesian product of all combinations of valid split points for both languages is tried,
and the pair of splits minimizing the cost function is selected,
unless not splitting results in even lower cost.

\section{Extracting cognates from parallel data}

Finnish--Estonian cognates were automatically extracted from the shared task training data.
As we needed a Finnish--Estonian parallel data set,
we generated one by triangulation from the English--Finnish and English--Estonian parallel data.
This resulted in a set of \numprint{679252} sentence pairs (ca 12 million tokens per language).

FastAlign \cite{fastalign} was used for word alignment in both directions,
after which the alignments were symmetrized using the \emph{grow-diag-final-and} heuristic.
All aligned word pairs were extracted based on the symmetrized alignment.
Words containing punctuation, and pairs aligned to each other fewer than 2 times were removed.
The list of word pairs was filtered based on Levenshtein distance.
If either of the words consisted of 4 or fewer characters, an exact match was required.
Otherwise, a Levenshtein distance up to a third of the mean of the lengths, rounding up, was allowed.
This procedure resulted in a list of \numprint{40472} cognate pairs.
The list contains words participating in multiple cognate pairs.
Cognate Morfessor is only able to link a word to a single cognate.
We filtered the list, keeping only the pairing to the most frequent cognate,
which reduces the list to \numprint{22226} pairs.

The word alignment provides a check for semantic similarity in the form of translational equivalence.
Even though the word alignment may produce some errors,
accidentally segmenting false friends consistently should not be problematic.

\section{Data}
After filtering, we have 9 million multilingual sentence pairs in total.
6.3M of this is English--Finnish,
of which 2.2M is parallel data,
and 4.1M is synthetic backtranslated data.
Of the 2.8M total English--Estonian,
1M is parallel
and 1.8M backtranslated.
The sentences backtranslated from Finnish were
from the news.2016.fi corpus,
translated with a PB-SMT model,
trained with WMT16 constrained settings.
The backtranslation from Estonian was freshly made
with a BPE-based system similar to our baseline system,
trained on the WMT18 data.
The sentences were selected from the news.20\{14-17\}.et corpora,
using a language model filtering technique.

\subsection{Preprocessing}

The preprocessing pipeline consisted of
filtering by length%
\footnote{1--100 tokens, 3--600 chars, $\leq$ 50 chars$/$token.}
and ratio of lengths%
\footnote{Requiring ratio 0.5--2.0, if either side $>$ 10 chars.},
fixing encoding problems,
normalizing punctuation,
removing of rare characters%
\footnote{$<10$ occurrences},
deduplication,
tokenizing,
truecasing,
rule-based filtering of noise,
normalization of contractions,
and filtering of noise using a language model.

The language model based noise filtering 
was performed by training a character-based deep LSTM language model on the in-domain monolingual data,
using it to score each target sentence in the parallel data,
and removal of sentences with perplexity per character above a manually picked threshold.
A lenient threshold%
\footnote{96\% of the data was retained.}
was selected in order to filter noise,
rather than for aiming for domain adaptation.
The same process was applied to filter the Estonian news data for backtranslation.

Our cognate segmentation resulted in a target vocabulary
of \numprint{42386} subwords for Estonian and \numprint{46930} subwords for Finnish,
resulting in \numprint{64396} subwords when combined.

For segmentation of the English source,
a separate Morfessor Baseline model was trained.
To ensure consistency between source and target segmentations,
we used the segmentation of the Cognate Morfessor model for any English words
that were also present in the target side corpora.
The source vocabulary consisted of \numprint{61644} subwords.

As a baseline segmentation,
we train a shared 100k subword vocabulary using BPE.
To produce a balanced multilingual segmentation, the following procedure was used:
First, word counts were calculated individually for
English and each of the target languages Finnish and Estonian.
The counts were normalized to equalize the sum of the counts for each language.
This avoided imbalance in the amount of data skewing the segmentation in favor of some language.
BPE was trained on the balanced counts.
Segmentation boundaries around hyphens were forced, overriding the BPE.

Multilingual translation with target-language tag
was done following \cite{johnson2016google}.
A pseudo-word, e.g. {\sc<to\_et>} to mark Estonian as the target language,
was prefixed to each paired English source sentence.

\setlength{\tabcolsep}{0.3em}
\begin{table}
\begin{center}
{\small
\begin{tabular}{rclr rclr rclr}
\toprule
$\epsilon$&$\rightarrow$&n   & 27919 &        g  &$\rightarrow$& k          & 3000  &         il & $\rightarrow$ & $\epsilon$  & 2077  \\
$\epsilon$&$\rightarrow$&a   & 17082 &        ü  &$\rightarrow$& y          & 2979  &          m & $\rightarrow$ & mm          & 2016  \\
$\epsilon$&$\rightarrow$&i   & 15725 &       oo  &$\rightarrow$& o          & 2790  &          s & $\rightarrow$ & n           & 2005  \\
       d  &$\rightarrow$&t   & 12599 &        t  &$\rightarrow$& a          & 2674  &         ee & $\rightarrow$ & e           & 1950  \\
       l  &$\rightarrow$&ll  & 5236  & $\epsilon$&$\rightarrow$& k          & 2583  &          i & $\rightarrow$ & $\epsilon$  & 1889  \\
$\epsilon$&$\rightarrow$&ä   & 4437  &       aa  &$\rightarrow$& a          & 2536  & $\epsilon$ & $\rightarrow$ & e           & 1803  \\
       s  &$\rightarrow$&ssa & 3907  &        õ  &$\rightarrow$& o          & 2493  &          u & $\rightarrow$ & o           & 1724  \\
       t  &$\rightarrow$&tt  & 3863  &        a  &$\rightarrow$& ä          & 2479  & $\epsilon$ & $\rightarrow$ & d           & 1496  \\
       o  &$\rightarrow$&u   & 3768  &        s  &$\rightarrow$& $\epsilon$ & 2173  &         il & $\rightarrow$ & t           & 1486  \\
       e  &$\rightarrow$&i   & 3182  &        t  &$\rightarrow$& $\epsilon$ & 2158  &          d & $\rightarrow$ & $\epsilon$  & 1433  \\
\bottomrule
\end{tabular}
}
\caption{30 most frequent edits learned by the model.
         The direction is Estonian$\rightarrow$Finnish.
         The numbers indicate how many times the edit was applied in the morph lexicon.
         $\epsilon$ indicates the empty string.
    \label{tab:edits}}
\end{center}
\end{table}
\setlength{\tabcolsep}{0.5em}

\section{NMT system}

We use the OpenNMT-py \cite{opennmt} implementation of the Transformer.

\subsection{Transformer}

The Transformer architecture \cite{vaswani2017attention}
relies fully on attention mechanisms, without need for recurrence or convolution.
A Transformer is a deep stack of layers, consisting of two types of sub-layer:
multi-head (MH) attention (Att) sub-layers and feed-forward (FF) sub-layers:
\begin{align}
\attention(Q, K, V) & = \softmax(\frac{QK^T}{\sqrt{d_{k}}})V \nonumber\\
a_{i}        &      = \attention(QW_{i}^{Q}, KW_{i}^{K}, VW_{i}^{V}) \nonumber\\
\multihead(Q, K, V) & = [a_{1};\mathellipsis;a_{h}]W^{O} \nonumber\\
\feedforward(x)     & = \max(0, xW_{1} + b_{1})W_{2} + b_{2}
\end{align}
where $Q$ is the input query,
$K$ is the key,
and $V$ the attended values.
Each sub-layer is individually wrapped in a residual connection and layer normalization.

When used in translation, Transformer layers are stacked into an encoder-decoder structure.
In the encoder, the layer consists of a self-attention sub-layer followed by a FF sub-layer.
In self-attention, the output of the previous layer is used as queries, keys and values $Q = K = V$.
In the decoder, a third context attention sub-layer is inserted between the self-attention and the FF.
In context attention, $Q$ is again the output of the previous layer,
but $K = V$ is the output of the encoder stack.
The decoder self-attention is also masked to prevent access to future information.
Sinusoidal position encoding makes word order information available.

\subsection{Training}

\begin{table}
\begin{center}
{\small
\begin{tabular}{lrr}
\toprule
                                & chrF-1.0         & BLEU\%           \\
\sc{en-et}                      & dev              & dev              \\
\midrule
BPE                             &     56.52        &     17.93           \\
monolingual                     &     53.44        &     15.82           \\
Cognate Morfessor               &     57.05        &     18.40           \\
\quad +finetuned                &     57.23        &     18.45           \\
\quad\quad +ensemble-of-5       & \bb 57.75        & \bb 19.09           \\
\quad\quad +ensemble-of-3       &     57.64        &     18.96           \\
\quad +linked embeddings        &     56.20        &     17.48           \\
\quad $-$LM filtering           &     52.94        &     14.65           \\
\quad 6+6 layers                &     57.35        &     18.84           \\
\bottomrule
\end{tabular}
}
\caption{Development set results for English--Estonian.
    character-F and BLEU scores in percentages.
    $+$/$-$ stands for adding/removing a component.
    Multiple modifications are indicated by increasing the indentation.
    \label{tab:etresults}}
\end{center}
\end{table}

\begin{table*}
\begin{center}
{\small
\begin{tabular}{lrrrrrrrr}
\toprule
                          & \multicolumn{4}{c}{chrF-1.0}                                  & \multicolumn{4}{c}{BLEU\%}                                               \\
\cmidrule(lr){2-5} \cmidrule(lr){6-9}
\sc{en-fi}                & nt2015        & nt2016       &  nt2017       &  nt2017AB      & nt2015         & nt2016         & nt2017          & nt2017AB     \\
\midrule
BPE                       & 58.59         & 59.76        &  62.00        &  63.06         & 21.09          & 21.04          & 23.49          & 26.55             \\
monolingual               & 57.94         & 59.11        &  61.33        &  62.41         & 20.87          & 20.70          & 23.11          & 26.12             \\
Cognate Morfessor         & 58.18         & 59.81        &  62.15        &  63.24         & 20.73          & 21.18          & 23.37          & 26.26             \\
\quad +finetuned          & 58.48         & 59.89        &  62.17        &  63.28         & 21.08          & 21.41          & 23.45          & 26.52             \\
\quad\quad +ensemble-of-8 & \bb 59.07     & \bb 60.69    &  \bb 62.94    &  \bb 64.07     & \bb 21.50      & \bb 22.34      & \bb 24.59      & \bb 27.55         \\
\quad $-$LM filtering     & 58.19         & 59.39        &  61.78        &  62.82         & 20.62          & 20.77          & 23.38          & 26.36             \\
\quad +linked embeddings  & 57.79         & 59.45        &  61.52        &  62.58         & 19.95          & 20.84          & 22.70          & 25.69             \\
\quad 6+6 layers          & 58.68         & 60.26        &  62.37        &  63.52         & 21.05          & 21.81          & 23.93          & 27.08             \\
\bottomrule
\end{tabular}
}
\caption{Results for English--Finnish.
    character-F and BLEU scores in percentages.
    $+$/$-$ stands for adding/removing a component.
    Newstest is abbreviated nt.
    Both references are used in nt2017AB.
    \label{tab:firesults}}
\end{center}
\end{table*}

Based on some preliminary results,
we decided to reduce the number of layers to 4 in both encoder and decoder;
later we found that the decision was based on too short training time.
Other parameters were chosen following the OpenNMT FAQ \cite{FAQ}:
512-dimensional word embeddings and hidden states,
dropout 0.1, batch size 4096 tokens, label smoothing 0.1,
Adam with initial learning rate 2 and $\beta_{2}$ 0.998.

Fine-tuning for each target language was performed
by continuing training of a multilingual model.
Only the appropriate monolingual subset of the training data was used in this phase.
The data was still prefixed for target language as during multilingual training.
No vocabulary pruning was performed.

In our ensemble decoding procedure,
the predictions of 3--8 models
are combined by averaging after the softmax layer.
Best results are achieved when the models have been independently trained.
However, we also try combinations where a second copy of a model is
further trained with a different configuration (monolingual finetuning).

We experimented with partially linking the embeddings of cognate morphs.
In this experiment, we used morph embeddings concatenated from two parts:
a part consisting of normal embedding of the morph,
and a part that was shared between both halves of the cognate morph pair.
Non-cognate morphs used an unlinked embedding also for the second part.
After concatenation, the linked embeddings have the same size as the baseline embeddings.

We evaluate the systems with cased BLEU using the mteval-v13a.pl script,
and characterF \cite{popovic2015chrf} with $\beta$ set to 1.0.
The latter was used for tuning.

\section{Results}

Based on preliminary experiments,
the Morfessor corpus cost weight $\alpha$ was set to 0.01,
and the edit cost weight was set to 10.
The most frequent edits are shown in Table~\ref{tab:edits}.

Table~\ref{tab:etresults} shows the development set results for Estonian.
Table~\ref{tab:firesults} shows results for previous year's test sets for Finnish.

The tables show our main system and the two baselines:
a multilingual model using joint BPE segmentation,
and a monolingual model using Morfessor Baseline.

Cognate Morfessor outperforms the comparable BPE system according to both measures
for Estonian, and according to chrF-1.0 for Finnish.
For Finnish, results measured with BLEU vary between test sets.
The cross-lingual segmentation is particularly beneficial for Estonian.

In the monolingual experiment,
the cross-lingual segmentations are replaced with monolingual Morfessor Baseline segmentation,
and only the data sets of one language pair at a time is used.
These results show that even the higher resourced language, Finnish, benefits from multilingual training.

The indented rows show variant configurations of our main system.
Monolingual finetuning consistently improves results for both languages.
For Estonian, we have two ensemble configurations:
one combining 3 monolingually finetuned independent runs,
and one combining 5 monolingually finetuned savepoints from 4 independent runs.
Selection of savepoints for the ensemble was based on development set chrF-1.
In the ensemble-of-5, one training run contributed two models:
starting finetuning from epochs 14 and 21 of the multi-lingual training.
The submitted system is the ensemble-of-3,
as the ensemble-of-5 finished training after the deadline.
For Finnish, we use an ensemble of 4 finetuned and 4 non-finetuned savepoints
from 4 independent runs.

To see if further cross-lingual learning could be achieved,
we performed an unsuccessful experiment with linked embeddings.
It appears that explicit linking does not improve the morph representations
over what the translation model is already capable of learning.

After the deadline, we trained a single model with 6 layers in both the encoder and decoder.
This configuration consistently improves results compared to the submitted system.

All the variant configurations (ensemble, finetuning, LM filtering, linked embeddings, number of layers)
used with Cognate Morfessor are compatible with each other.
We did not not explore the combinations in this work,
except for combining finetuning with ensembleing: 
all of the models in the Estonian ensembles, and 4 of the models in the Finnish ensemble are finetuned.
All the variant configurations except for linked embeddings could also be used with BPE.

\section{Conclusions and future work}

The translation system trained using the Cognate Morfessor segmentation
outperforms the baselines for both languages.
The benefit is larger for Estonian, the language with less data in this experiment.

One downside is that, due to the model structure, 
Cognate Morfessor is currently not applicable to more than two target languages.

Cognate Morfessor itself learns to model the frequent edits between cognate pairs.
However, in the preprocessing cognate extraction step of this work,
we used unweighted Levenshtein distance, which does not distinguish edits by frequency.
In future work, weighted or graphonological Levenshtein distance could be applied \cite{babych2016graphonological}.

\section*{Acknowledgments}
This research has been supported by
the European Union's Horizon 2020 Research and Innovation Programme under Grant Agreement No 780069. 
Computer resources within the Aalto University School of Science ``Science-IT'' project were used.
We wish to thank Peter Smit for groundlaying work that led to Cognate Morfessor.

\bibliographystyle{acl_natbib_nourl}
\bibliography{2018_wmt_cognate}

\end{document}

%% file: core.tex
\usepackage{amsmath}
\usepackage{ifthen}
\usepackage{booktabs}   
\usepackage{microtype}  

\DeclareMathOperator* \cost {L}

\newcommand{\vb}{\,|\,}

\newcommand{\params}{\boldsymbol{\theta}}
\newcommand{\data}{\boldsymbol{D}}

\newcommand{\fixme}[2][]{%
    \ifthenelse{\equal{#1}{}}%
               {\textsl{[#2]}}%
               {\textsl{[#1: #2]}}%
    }%

\newcommand{\examp}[1]{\emph{#1}}

\usepackage{numprint}
\npdecimalsign{.}
\npthousandsep{\ensuremath{\;}}
\npfourdigitnosep

\newcommand{\bb}{\fontseries{b}\selectfont}